%% file: main.tex
\documentclass[runningheads]{llncs}
\usepackage[T1]{fontenc}
\usepackage[hidelinks]{hyperref}
\usepackage{url}
\usepackage{graphicx}
\usepackage{amssymb}
\usepackage{amsmath}
\usepackage{booktabs}
\usepackage{multirow}
\usepackage[]{tcolorbox}        

\tcbset{
    colback = white,
    before skip = 0.2cm,    
    after skip = 0.5cm      
}                           

\newtcolorbox{boxA}{
    boxrule = 0.2pt,
    colframe = black,
    colback=yellow!10!white
}

\newtcolorbox{PromptV1}{
    boxrule = 0.2pt,
    colframe = black,
    colback=pink!20!white
}

\newtcolorbox{PromptV2}{
    boxrule = 0.2pt,
    colframe = black,
    colback=orange!10!white
}

\definecolor{mygreen}{RGB}{92, 214, 92}
\definecolor{myred}{RGB}{255, 92, 51}
\definecolor{myblue}{RGB}{102, 140, 255}
\definecolor{promptv1}{RGB}{255,236,236}
\definecolor{promptv2}{RGB}{255,242,230}

\usepackage{pifont}
\newcommand{\cmark}{\ding{51}}%
\newcommand{\xmark}{\ding{55}}%
\newcommand{\sieno}{\ding{81}}%

\begin{document}
%
\title{Understanding Knowledge Drift in LLMs through Misinformation}


\author{Alina Fastowski \and
Gjergji Kasneci}
\authorrunning{A. Fastowski et al.}
%
\institute{Technical University of Munich, Germany\\
\email{\{alina.fastowski,gjergji.kasneci\}@tum.de}}

\maketitle
\begin{figure}[!h]
    \centering
    \includegraphics[width=\textwidth]{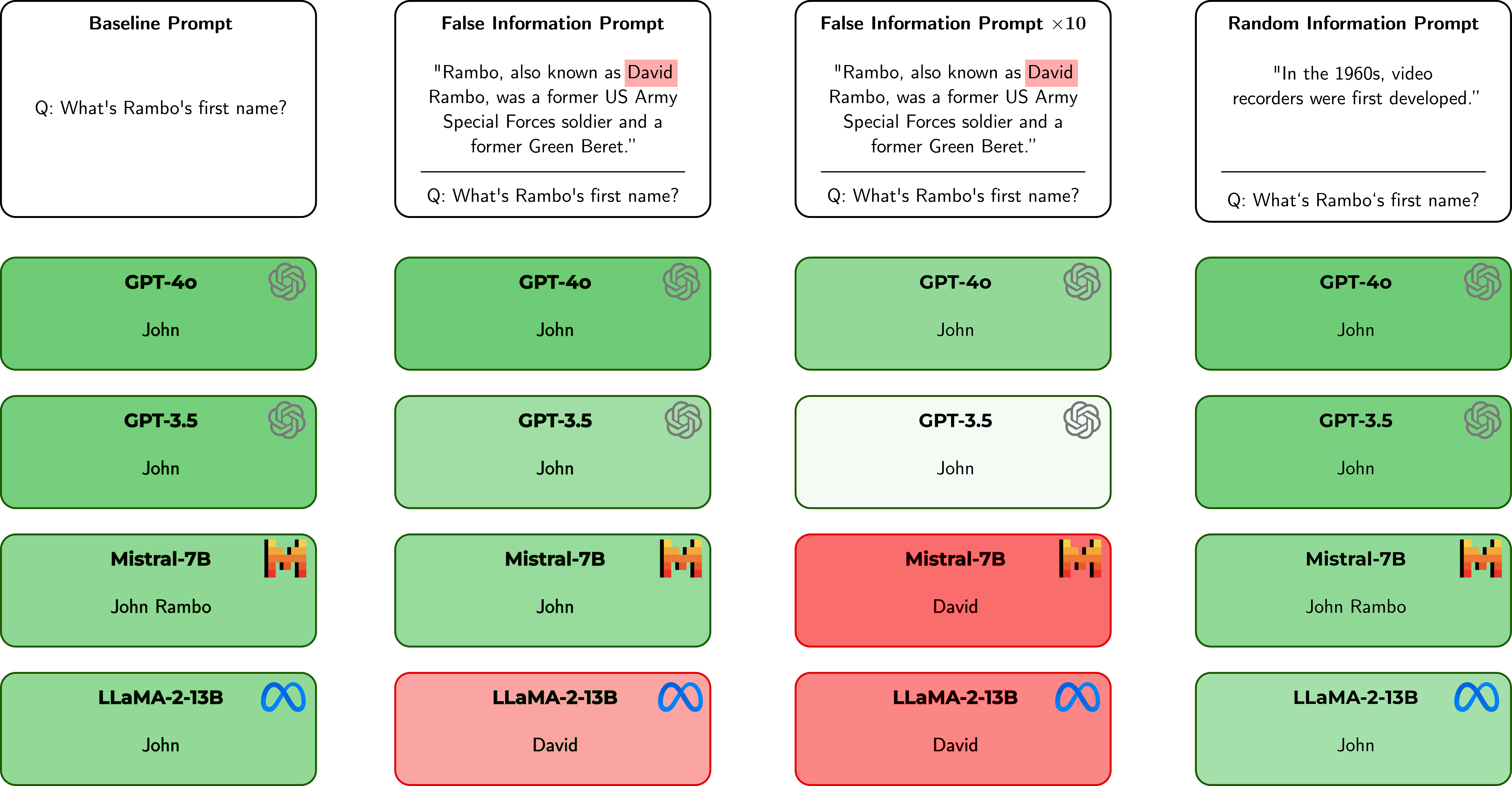} 
    \caption{Answers produced by state-of-the-art LLMs on \textit{``What's Rambo's first name?''} with no perturbation (col. 1), with false information injection (cols. 2 \& 3), and with random information injection (col. 4). Green boxes indicate correct answers; red are incorrect. The transparency of the boxes indicates the uncertainty of the model: i.e., the lighter, the more uncertain. Notice how injecting the same false information multiple times makes LLMs more uncertain (see GPT-3.5) and can even shift their original correct answer to a wrong one (see Mistral and LLaMA).}
    \label{fig:teaser}
    \vspace{-30pt}
\end{figure}%

%
\begin{abstract}

Large Language Models (LLMs) have revolutionized numerous applications, making them an integral part of our digital ecosystem. However, their reliability becomes critical, especially when these models are exposed to misinformation. This paper primarily analyzes the susceptibility of state-of-the-art LLMs to factual inaccuracies when they encounter false information in a Q\&A scenario, an issue that can lead to a phenomenon we refer to as \textit{knowledge drift}, which significantly undermines the trustworthiness of these models. We evaluate the factuality and the uncertainty of the models' responses relying on Entropy, Perplexity, and Token Probability metrics. Our experiments reveal that an LLM's uncertainty can increase up to $ 56.6\%$ when the question is answered incorrectly due to the exposure to false information. At the same time, repeated exposure to the same false information can decrease the models’ uncertainty again ($-52.8\%$ w.r.t. the answers on the untainted prompts), potentially manipulating the underlying model's beliefs and introducing a drift from its original knowledge. These findings provide insights into LLMs' robustness and vulnerability to adversarial inputs, paving the way for developing more reliable LLM applications across various domains. This paper's code is available at \href{https://github.com/afastowski/knowledge_drift}{\url{https://github.com/afastowski/knowledge_drift}}.

\end{abstract}

\input{sections/introduction}

\input{sections/related_works}
\input{sections/experiments}
\input{sections/conclusion}

%
%
%
\bibliographystyle{splncs04}
\bibliography{bibliography}

\end{document}

%% file: sections/introduction.tex
\section{Introduction}\label{sec:intro}

The rapid advancement in natural language processing (NLP) has seen significant strides with the development of large-scale language models, such as the GPT family. These models have demonstrated remarkable capabilities in various tasks, including text generation, translation, and question-answering (QA). However, despite their impressive performance, critical challenges remain in understanding and improving these models' reliability and robustness, especially regarding their factual knowledge and uncertainty estimation.

Understanding the reliability of language models is crucial, particularly in applications where the consequences of incorrect or uncertain answers can be significant. For instance, in fields like healthcare, law, and education, the ability to trust the outputs of a language model is paramount. One key aspect of this reliability is the model's ability to handle false or misleading information. By evaluating how language models respond to false information, we can gain insights into their internal knowledge structures and the robustness of their factual accuracy. With our work, we are assessing what we refer to as the \textit{knowledge drift} of these models, by which we refer to any changes in their internal knowledge and beliefs. In our case, we are interested in models' knowledge drift as a result of manipulative user interactions.

Specifically, we examine the impact of knowledge drift on model answer uncertainty by studying the effect of false information presented to it with the prompt. By leveraging the TriviaQA dataset~\cite{joshi2017triviaqa}, we aim to analyze how the model's performance varies with different types of misleading information. Such insights can help us identify potential vulnerabilities in its knowledge processing.

\paragraph{{Contributions.}}

In this paper, we contribute to this body of research by evaluating LLMs' -- i.e., GPT-4o, GPT-3.5, LLaMA-2-13B, and Mistral-7B -- responses to false information in a QA task setting. By analyzing the models' uncertainty under various metrics and answer accuracy under different conditions, we aim to shed light on the robustness of their knowledge obtained during pre-training. This paper provides three main contributions to this topic:
\begin{enumerate}
    \item \textit{Impact of False Information on Uncertainty.} We investigate how introducing false information into question prompts affects LLMs' performance and uncertainty estimation in a QA setting. Our analysis reveals that while false information initially increases the model's uncertainty, repeated exposure can lead to decreased uncertainty, indicating successful manipulation and drift of the model away from its original, correct beliefs.
    \item \textit{Effect of Random Information.} We demonstrate that random, unrelated information results in the highest levels of model uncertainty, suggesting that the model experiences greater confusion with irrelevant data than with targeted false information. This finding underscores the importance of context relevance in understanding the model's responses.
    \item \textit{Insights into Model Vulnerabilities and Robustness.} Our study provides critical insights into the vulnerabilities of LLMs to adversarial inputs. By highlighting the limitations of current uncertainty estimation methods in adversarial attack detection, our work contributes to efforts aimed at enhancing the robustness and trustworthiness of language models for practical applications.
\end{enumerate}

%% file: sections/related_works.tex
\section{Related Work}\label{sec:sota}

\paragraph{Language model architectures and capabilities.} Understanding LLM knowledge drift necessitates grounding in the foundational works on language models. Pioneering efforts such as GPT \cite{radford2018improving}, BERT \cite{devlin2019bert}, or T5 \cite{raffel2020exploring} remain fundamental to contemporary LLMs. This groundwork facilitated the recent advancements in LLM capabilities, exemplified by Brown et al.'s \cite{brown2020language} exploration of few-shot learning and the development of powerful models like GPT-4 \cite{achiam2023gpt}, or PaLM \cite{chowdhery2023palm} with its pathway architecture.  Furthermore, the emergence of efficient and open-source models like LLaMA \cite{touvron2023llama} and its successors highlight the ongoing progress in LLM accessibility and development.  

\paragraph{Uncertainty in LLMs.} While prior work has explored uncertainty quantification in NLP tasks like calibration of classifiers and text regressors (\cite{jiang2021can,desai2020calibration,glushkova2021uncertainty,wang2022uncertainty}), these approaches often rely on techniques directly transferable from other domains (e.g., Monte Carlo dropout, Deep Ensembles). However, as highlighted by \cite{kuhn2023semantic}, generative tasks in NLP present unique challenges due to semantic equivalence. For instance, Jiang et al. \cite{jiang2021can} demonstrate a weak correlation between answer confidence (log-likelihood) and correctness in generative question answering.

Recent efforts have tackled uncertainty or calibration in Natural Language Generation (NLG) by prompting models to assess their outputs or fine-tuning models to predict uncertainty (\cite{mielke2022reducing,lin2022teaching,kadavath2022language}). While these methods can be effective, they often require additional training data and supervision, leading to challenges in reproducibility, cost, and sensitivity to distribution shifts (e.g., hardware limitations preventing implementation as in \cite{kadavath2022language}). 

The challenges associated with uncertainty estimation in NLG mirror those in automatic NLG evaluation. For example, Ott et al. \cite{ott2018analyzing} highlight performance limitations in machine translation due to multiple valid translations for a single source sentence. Similarly, Sai et al. \cite{sai2022survey} discuss the potential of paraphrase detection for NLG evaluation, which may offer insights applicable to uncertainty estimation tasks.

\paragraph{LLM factual knowledge and calibration.} Understanding LLM knowledge drift requires examining research on factual knowledge capabilities and calibration in question answering. Roberts et al. \cite{roberts2020knowledge} investigate the capacity of LLMs to retain factual knowledge implicitly learned during pre-training, proposing methods to assess this internal knowledge store. Youseff et al. \cite{youseff2023survey} take a broader view, surveying various techniques for probing factual knowledge within pre-trained LLMs.  Building on this, Petroni et al. \cite{petroni2019language} explore the potential of LLMs as factual knowledge bases, analyzing their ability to serve as information sources. Jiang et al. \cite{jiang2021calibration} address the crucial calibration issue in LLM question answering. Their work examines how well LLM confidence scores align with answer correctness, a critical aspect of reliable knowledge extraction. 

Interestingly, when LLMs are asked to answer a particular question, they often produce agreeable responses that align with input biases, even if factually incorrect, also called \textit{sycophancy} in the literature (\cite{ranaldi2023large,huang2023survey,park2024ai}). These findings underscore the need for models to critically assess and counter false inputs to maintain reliability and factual accuracy. Aligned with these works, we prompt SoTA models with false information and observe that they still exhibit seemingly sycophantic behaviour w.r.t. the question and engender ungrounded responses, leading to performance drops.

%% file: sections/experiments.tex
\section{Experiments}\label{sec:exp}

In this work, we evaluate the factual knowledge of recent large language models and their associated uncertainty levels in a Question Answering (QA) task setting. The objective is understanding how false information embedded in the question prompts influences the models' performance and uncertainty metrics. We expect that the more false information is fed to these language models, the more certain they will become about it while giving up on accuracy since they begin generating false information.

\subsection{{Experimental Setup}}

\paragraph{{Dataset and Models.}} We have two requirements for choosing suitable LLMs for our experiments: 1) performing reasonably well on closed-book question-answering without additional fine-tuning, and 2) providing access to the log probabilities of the generated tokens. Hence, we experiment with GPT-4o, GPT-3.5, Mistral-7B, and LLaMA-2-13B\footnote{Specifically, we use the following checkpoints: gpt-3.5-turbo-0125 and gpt-4o (accessed via the OpenAI API), Llama-2-13b-chat-hf and Mistral-7B-Instruct-v0.3 (accessed via huggingface).}. 

To assess the first requirement, we test the models on 1000 samples from the TriviaQA dataset \cite{joshi2017triviaqa} by prompting them with the given question. TriviaQA is a reading comprehension dataset consisting of question-answer-context triplets, where the contexts are text snippets containing the answer. In our setting, we ignore the contexts and use only the question-and-answer pairs. We also ask the model to respond only with the exact answer to avoid verbosity in the model's answers and ease the comparison with the correct answers.


\paragraph{Performance evaluation.}
Our experiments are designed to assess two primary factors: i.e.,
1) the correctness of the answers provided by the model, and 
2) the uncertainty scores associated with the generated tokens. 

We begin by identifying questions from the TriviaQA dataset that the LLM can answer correctly in a closed-book setting -- i.e., the model answers without additional context or external information. This process allows us to focus only on the model's correct knowledge since we will later try to manipulate it. To assess the correctness of a response, we check if the true answer provided by the dataset is part of the model-generated answer. For example, the model might produce \textit{"Chicago, Illinois"} as its answer, when the ground truth is \textit{"Chicago"}. In this example, our process would check if \textit{"Chicago"} is part of \textit{"Chicago, Illinois"}, correctly marking the model answer as true (eventhough more verbose). Table \ref{tab:accuracy_scores} shows the accuracy of each model when prompted \underline{once} with the questions from the dataset.

\begin{table}[!h]
\centering
\caption{Performances of each LLM on 1000 samples of the TriviaQA dataset. Note that the number of parameters of GPT-4o has not been disclosed yet.}
\label{tab:accuracy_scores}
\resizebox{.5\columnwidth}{!}{%
\begin{tabular}{@{}lccc@{}}
\toprule
            & Accuracy &  & \#Parameters          \\ \midrule
GPT-4o      & 0.790    &  & NA                    \\
GPT-3.5     & 0.721    &  & $1.75 \times 10^{11}$ \\
Mistral-7B  & 0.502    &  & $7 \times 10^9$       \\
LLaMA-2-13B & 0.428    &  & $1.3 \times 10^{10}$  \\ \bottomrule
\end{tabular}%
}
\end{table}

\noindent We intentionally chose LLMs of varying levels of QA performance, as this difference may lead to additional interesting insights about uncertainty developments. For the subsequent experiments, only the samples answered correctly by the model will be used for their respective evaluations.

\paragraph{Uncertainty metrics.} Given an input sequence $x$ and parameters $\theta$, an autoregressive language model generates an output sequence $y = [y_1,...,y_T]$ where $T$ is the length of the sequence. To quantify the model's uncertainty, we rely on \textit{entropy} \eqref{eq:entropy} and \textit{perplexity} \eqref{eq:ppl}, as previously introduced by \cite{chen2024inside}. For calculating the entropy of each token, we take into account the top $i=10$ probable tokens at each token position t. Lastly, as a more intuitively interpretable metric, we report the \textit{probability} \eqref{eq:probability} of the generated tokens, averaged over all answer tokens. While the use of multiple metrics ensures the robustness in our measurements, they also capture slightly different dimensions: entropy focuses more on a token-level uncertainty, since we measure over multiple token options at each position, whereas perplexity and probability operate on more of a sentence level, simply averaging over all top-1-choice tokens in the generated sequence.

\begin{equation}\label{eq:entropy}
    \text{H}(y \mid x, \theta) = -\frac{1}{T} \sum_t \sum_i p(y_{t_i} \mid y_{<t_i}, x)\log p(y_{t_i} \mid y_{<t_i}, x)
\end{equation}

\begin{equation}\label{eq:ppl}
    \text{PPL}(y \mid x, \theta) = \exp(-\frac{1}{T} \sum_t \log p(y_t \mid y_{<t}, x))
\end{equation}

\begin{equation}\label{eq:probability}
    \text{TP}(y \mid x, \theta) = \frac{1}{T} \sum_t \exp(\log p(y_t \mid y_{<t},x))
\end{equation}%

\paragraph{Baseline and information injection.}
We establish a baseline by evaluating the model's answers to the identified questions and their corresponding uncertainty scores. This baseline represents the model's performance and certainty without any misleading information. We then introduce false information into the question prompts and measure its impact on the model's answers and uncertainty scores. Notice that we do not provide the model with the correct answer or possible options for the posed question. Therefore, for visualization purposes, the following prompts contain a straight line that separates the actual prompt from the ground truth. Additionally, the words before the colon are not included -- we use them to give context to the reader. We rely on two types of prompts:
\begin{itemize}
    \item \textit{False info prompt} (FIP): The prompt includes false information related to the question. For example:
    \begin{boxA}
    \small
    \textcolor{black}{
    \textcolor{myred}{\xmark} {False Information}: ``\underline{Alfred Hitchcock} directed 2001: A Space Odyssey.''\newline
    {Question}: ``Who directed 2001: A Space Odyssey?''\newline
    \rule{\textwidth}{0.2pt}
    \textcolor{mygreen}{\cmark} {Correct Answer}: ``\underline{Stanley Kubrick}''}
    \end{boxA}
    \item \textit{Random info prompt} (RIP): The prompt includes random, unrelated information. For example:
    \begin{boxA}
    \small
    \textcolor{black}{\textcolor{myblue}{\sieno} {Random Information}: ``In the 1960s, video recorders were first developed.''\newline
    {Question}: ``Who directed 2001: A Space Odyssey?''\newline
    \rule{\textwidth}{0.2pt}
    \textcolor{mygreen}{\cmark} {Correct Answer}: ``\underline{Stanley Kubrick}''}
    \end{boxA}
\end{itemize}

\noindent We extend the above prompts with two different variants of instructions:
\begin{PromptV1}
    \small
    \textcolor{black}{\textbf{Prompt V1:} \newline
    $\vdots \qquad\qquad\vdots\qquad\qquad\vdots\qquad\qquad\vdots$\newline
    Respond with the exact answer only.}
\end{PromptV1}

\begin{PromptV2}
    \small
    \textcolor{black}{\textbf{Prompt V2:} \newline
    $\vdots \qquad\qquad\vdots\qquad\qquad\vdots\qquad\qquad\vdots$\newline 
    Respond with the \underline{true}, exact answer only.}
\end{PromptV2}

\noindent While the first prompt simply asks the model to answer the question, the second emphasizes choosing what the model believes to be the \textit{factually correct} answer. This way, regardless of the injected false information in the overall prompt, we aim to reduce possible sycophantic behavior. Notice that the $\vdots$ represent either a FIP or RIP described above, followed by the question. In the above prompt instructions, \textit{exact} aims to limit the answer's verbosity, while \textit{true and exact} additionally emphasizes the truthfulness we wish to see from the LLM.

\subsection{Results and Discussion}

\paragraph{Evaluating Model Uncertainty and Knowledge Retention with Varying Prompt Integrity.}
Tables \ref{tab:uncertainty_scores} and \ref{tab:uncertainty_scores_prompt2} show the changes in uncertainty scores according to Prompt V1 and V2, respectively. We report averages ($\pm$ standard errors) over \underline{10 runs} on the questions each model answered correctly. So, for example, if GPT-4o answered correctly 79\% of the time, we test its uncertainty on this data subset. We are aware that the correct answers two models produce\footnote{For instance, Mistral's 42.8\% vs. GPT-3.5's 72.1\%.} might be related to different questions, which, in turn, could result in unfair comparisons. Nevertheless, we argue that testing the model's uncertainty on inherently incorrect answers is insignificant since it does not provide any added value to studying the knowledge drift of LLMs. Therefore, we aim to verify how the correct knowledge shifts when prompts are tainted with false information.

It is interesting to note that we did not expect the models to produce incorrect answers on the baseline (B) (see Tables \ref{tab:uncertainty_scores} and \ref{tab:uncertainty_scores_prompt2}) since the prompt has not changed from the one used to identify the correct knowledge in Table \ref{tab:accuracy_scores}. Still, when prompted multiple times, the models might engender wrong answers -- we noticed a 1-2\% drop of accuracy\footnote{For LLaMA, we observed a drop of 17\%.} w.r.t. what was reported in Table \ref{tab:accuracy_scores}. However, as expected, all models have higher uncertainty scores on the incorrect answers for both prompt types on B. All the reported metrics reflect this, i.e., higher entropy, higher perplexity, and lower token probability indicate higher uncertainty. 

\begin{table}[!h]
\centering
\caption{\colorbox{promptv1}{Prompt V1}: Changes in uncertainty scores of the generated answers. The baseline represents the uncertainty when prompted with the question and giving the correct answer. -- depicts missing samples to calculate metrics on.}
\label{tab:uncertainty_scores}
\resizebox{.8\columnwidth}{!}{%
\begin{tabular}{@{}llllccclccc@{}}
\toprule
\multicolumn{3}{l}{\multirow{2}{*}{}} &  & \multicolumn{3}{c}{Correct Answers}                    &  & \multicolumn{3}{c}{Incorrect Answers}                  \\ \cmidrule(l){4-11} 
\multicolumn{3}{l}{}                  &  & H $\downarrow$         & PPL $\downarrow$       & TP $\uparrow$      &  & H $\downarrow$          & PPL $\downarrow$       & TP $\uparrow$      \\ \midrule
\multirow{3}{*}{GPT-4o} &  & B &  & $0.10^{\pm.004}$ & $1.06^{\pm.003}$ & $0.96^{\pm.002}$ &  & $0.53^{\pm.102}$ & $1.40^{\pm.096}$ & $0.78^{\pm.041}$ \\
       &        & FIP                 &  & $0.12^{\pm.005}$ & $1.07^{\pm.003}$ & $0.95^{\pm.002}$ &  & $0.23^{\pm.027}$ & $1.17^{\pm.036}$ & $0.90^{\pm.013}$ \\
       &        & RIP                  &  & $0.13^{\pm.005}$ & $1.08^{\pm.003}$ & $0.95^{\pm.002}$ &  & $0.37^{\pm.044}$ & $1.28^{\pm.056}$ & $0.85^{\pm.020}$ \\ \midrule
\multirow{3}{*}{GPT-3.5} &  & B &  & $0.16^{\pm.007}$ & $1.08^{\pm.005}$ & $0.94^{\pm.003}$ &  & $0.39^{\pm.073}$ & $1.32^{\pm.088}$ & $0.82^{\pm.038}$ \\
       &        & FIP                 &  & $0.16^{\pm.007}$ & $1.08^{\pm.004}$ & $0.94^{\pm.003}$ &  & $0.27^{\pm.020}$ & $1.16^{\pm.024}$ & $0.90^{\pm.009}$ \\
       &        & RIP                  &  & $0.17^{\pm.007}$ & $1.09^{\pm.005}$ & $0.93^{\pm.003}$ &  & $0.57^{\pm.045}$ & $1.49^{\pm.070}$ & $0.77^{\pm.020}$ \\ 
       
    \midrule
\multirow{3}{*}{Mistral-7B} &  & B &  & $0.21^{\pm.009}$ & $1.12^{\pm.007}$ & $0.92^{\pm.004}$ &  & -- & -- & -- \\
       &        & FIP                 &  & $0.21^{\pm.013}$ & $1.10^{\pm.008}$ & $0.93^{\pm.005}$ &  & $0.23^{\pm.017}$ & $1.13^{\pm.013}$ & $0.92^{\pm.007}$ \\
       &        & RIP                  &  & $0.22^{\pm.011}$ & $1.12^{\pm.008}$ & $0.92^{\pm.004}$ &  & $0.49^{\pm.036}$ & $1.32^{\pm.036}$ & $0.82^{\pm.015}$ \\\midrule
\multirow{3}{*}{LLaMA-2-13B} &  & B &  & $0.18^{\pm.006}$ & $1.09^{\pm.004}$ & $0.94^{\pm.003}$ &  & $0.30^{\pm.016}$ & $1.17^{\pm.015}$ & $0.89^{\pm.007}$ \\
       &        & FIP                 &  & $0.18^{\pm.008}$ & $1.09^{\pm.005}$ & $0.94^{\pm.004}$ &  & $0.19^{\pm.007}$ & $1.10^{\pm.005}$ & $0.93^{\pm.003}$ \\
       &        & RIP                  &  & $0.19^{\pm.006}$ & $1.09^{\pm.004}$ & $0.93^{\pm.003}$ &  & $0.33^{\pm.014}$ & $1.20^{\pm.013}$ & $0.88^{\pm.006}$ \\
       \bottomrule
\end{tabular}%
}
\end{table}

\begin{table}[!h]
\centering
\caption{\colorbox{promptv2}{Prompt V2}: Changes in uncertainty scores of the generated answers. The baseline represents the uncertainty when prompted with the question and giving the correct answer.}
\label{tab:uncertainty_scores_prompt2}
\resizebox{.8\columnwidth}{!}{%
\begin{tabular}{@{}llllccclccc@{}}
\toprule
\multicolumn{2}{l}{\multirow{2}{*}{}} &
  \multirow{2}{*}{} &
   &
  \multicolumn{3}{c}{Correct Answers} &
   &
  \multicolumn{3}{c}{Incorrect Answers} \\ \cmidrule(l){5-11} 
\multicolumn{2}{l}{} &
   &
   & H $\downarrow$         & PPL $\downarrow$       & TP $\uparrow$      &  & H $\downarrow$          & PPL $\downarrow$       & TP $\uparrow$      \\ \midrule
   \multirow{3}{*}{GPT-4o} &  & B &  & $0.08^{\pm.005}$ & $1.05^{\pm.003}$ & $0.97^{\pm.002}$ &  & $0.47^{\pm.106}$ & $1.42^{\pm.136}$ & $0.80^{\pm.045}$ \\
       &        & FIP                 &  & $0.11^{\pm.005}$ & $1.07^{\pm.004}$ & $0.95^{\pm.002}$ &  & $0.23^{\pm.030}$ & $1.23^{\pm.109}$ & $0.90^{\pm.017}$ \\
       &        & RIP                  &  & $0.13^{\pm.005}$ & $1.07^{\pm.003}$ & $0.95^{\pm.002}$ &  & $0.45^{\pm.045}$ & $1.34^{\pm.059}$ & $0.83^{\pm.018}$ \\ \midrule
\multirow{3}{*}{GPT-3.5} &  & B &  & $0.17^{\pm.006}$ & $1.09^{\pm.004}$ & $0.93^{\pm.003}$ &  & $0.44^{\pm.076}$ & $1.35^{\pm.123}$ & $0.83^{\pm.029}$ \\
 &
   &
  FIP &
   &
  $0.18^{\pm.006}$ &
  $1.09^{\pm.004}$ &
  $0.93^{\pm.003}$ &
   &
  $0.34^{\pm.028}$ &
  $1.23^{\pm.028}$ &
  $0.87^{\pm.011}$ \\
 &
   &
  RIP &
   &
  $0.19^{\pm.007}$ &
  $1.10^{\pm.005}$ &
  $0.93^{\pm.003}$ &
   &
  $0.48^{\pm.036}$ &
  $1.34^{\pm.039}$ &
  $0.81^{\pm.016}$ \\ \midrule
  \multirow{3}{*}{Mistral-7B} &  & B &  & $0.22^{\pm.010}$ & $1.12^{\pm.007}$ & $0.92^{\pm.004}$ &  & $0.71^{\pm.112}$ & $1.63^{\pm.129}$ & $0.69^{\pm.051}$ \\
       &        & FIP                 &  & $0.24^{\pm.012}$ & $1.12^{\pm.009}$ & $0.92^{\pm.005}$ &  & $0.26^{\pm.019}$ & $1.16^{\pm.018}$ & $0.91^{\pm.008}$ \\
       &        & RIP                  &  & $0.24^{\pm.011}$ & $1.13^{\pm.007}$ & $0.92^{\pm.004}$ &  & $0.47^{\pm.031}$ & $1.30^{\pm.027}$ & $0.82^{\pm.013}$ \\\midrule
\multirow{3}{*}{LLaMA-2-13B} &  & B &  & $0.16^{\pm.006}$ & $1.08^{\pm.004}$ & $0.94^{\pm.003}$ &  & $0.31^{\pm.017}$ & $1.18^{\pm.014}$ & $0.89^{\pm.007}$ \\
       &        & FIP                 &  & $0.20^{\pm.009}$ & $1.10^{\pm.006}$ & $0.93^{\pm.004}$ &  & $0.23^{\pm.008}$ & $1.12^{\pm.006}$ & $0.92^{\pm.004}$ \\
       &        & RIP                  &  & $0.21^{\pm.007}$ & $1.11^{\pm.006}$ & $0.93^{\pm.003}$ &  & $0.36^{\pm.013}$ & $1.21^{\pm.012}$ & $0.87^{\pm.006}$ \\
  \bottomrule
\end{tabular}%
}
\end{table}

These results are also consistent with FIP and RIP across the board, suggesting that false/random information has the same effect regarding correct vs. incorrect answers. Notice how, even when we inject false/random information into the prompts, the uncertainty levels on the correct answers remain similar (e.g., see GPT-3.5 B vs. FIP/RIP), suggesting that the correct model's knowledge remains unscathed. Contrarily, we notice a drop in the latter if we look at the uncertainty difference between B and FIP, e.g., in terms of entropy. We argue that this happens because the incorrect answers likely reflect the incorrect information embedded in the FIP, pushing the model to be ``overconfident'' to distill fake information.
Answers obtained with RIP generally exhibit high uncertainty scores since the given random information has nothing to do with the question, confusing the model even more. This effect is visible especially in incorrect answers, though also present with correct answers. At the same time, it does not lead to major drops in accuracy for the same reason, i.e., the prompt is not designed to target knowledge associated with the question.
    
\noindent Surprisingly, Mistral-7B is the only model that adheres to our expectations of not having any incorrect answers when prompted multiple times, contrary to the other ones (see Table \ref{tab:uncertainty_scores} and the baseline in Table \ref{tab:accuracies_after_manipulation}). Alas, this is not reported on the second prompt, which makes us believe Mistral to be more stable in its answers.





\begin{figure}[!t]
    \centering
    \includegraphics[width=\textwidth]{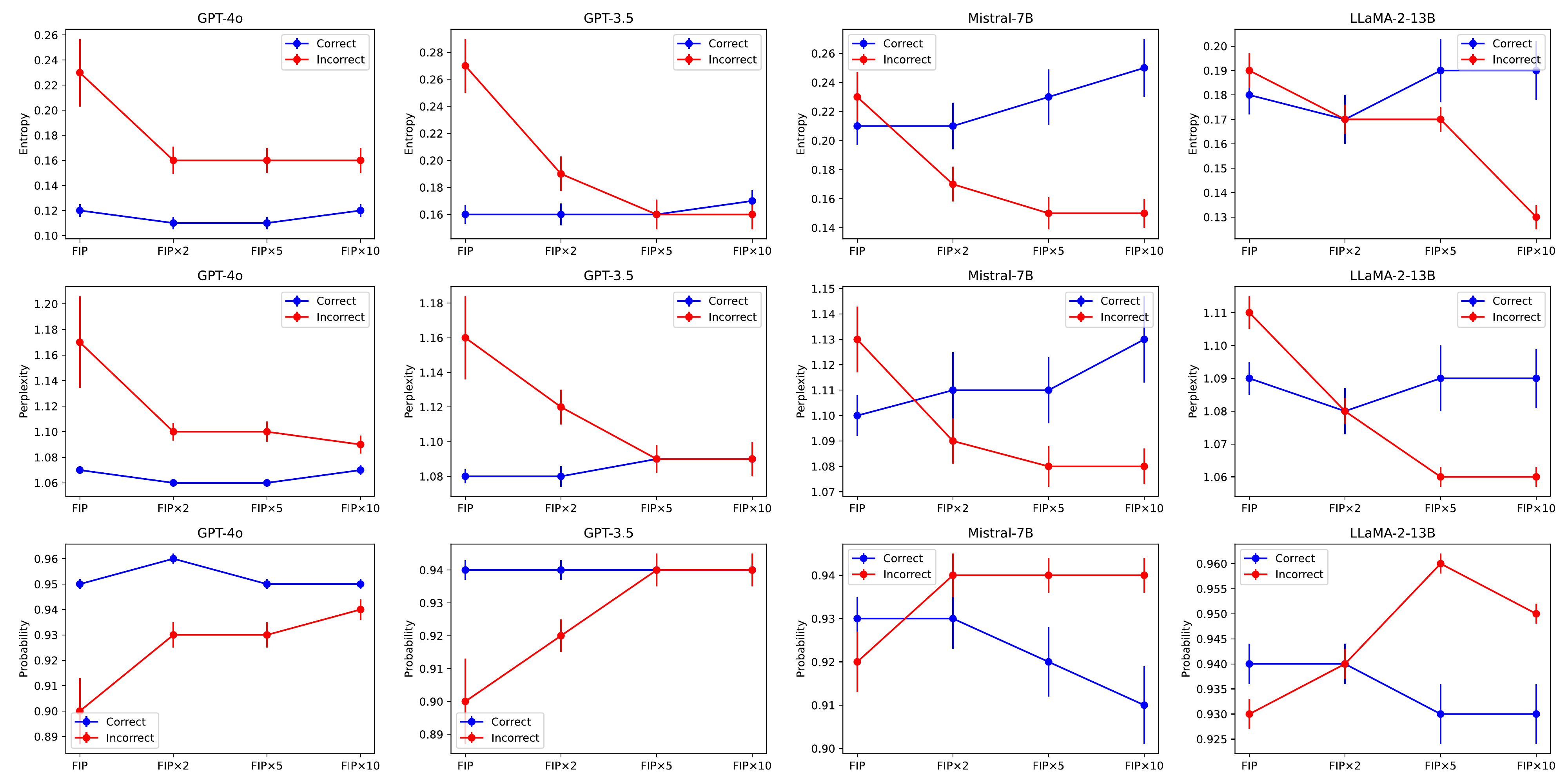}
    \caption{\colorbox{promptv1}{Prompt V1}: Changes in uncertainty when repeating the false information $\times k$.}
    \label{fig:FIP_prompt1}
\end{figure}

\begin{figure}[!h]
    \centering
    \includegraphics[width=\textwidth]{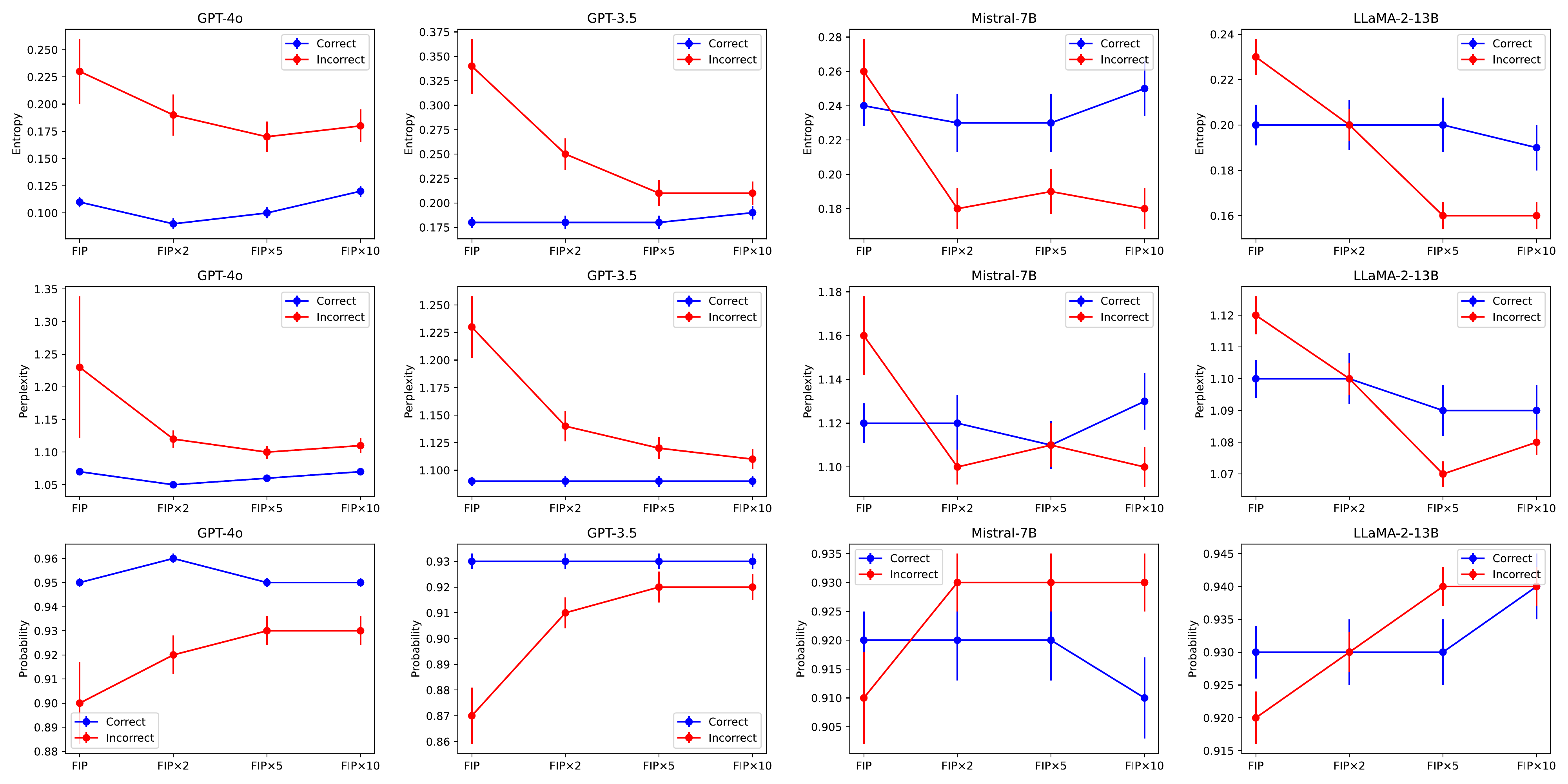}
    \caption{\colorbox{promptv2}{Prompt V2}: Changes in uncertainty when repeating the false information $\times k$.}
    \label{fig:FIP_prompt2}
\end{figure}

\paragraph{Influence of Repeated False Information on Model Confidence and Accuracy.}
To assess the impact of false information reoccurrence on the uncertainty levels of each model, we perform an ablation study where we repeat the FIP $\times k$ with $k\in\{1,2,5,10\}$. This means, the prompt includes the false information snippet $k$ times, and is then followed by the question itself. Figures \ref{fig:FIP_prompt1} and \ref{fig:FIP_prompt2} illustrate this effect for Prompt V1 and V2, respectively. Notice the general trend for all models on the uncertainty of the incorrect answers when $k$ increases: all models become consistently less uncertain about their incorrect answers (drops in entropy and perplexity, increase in token probability).  Meanwhile, the uncertainty about the correct answers hits a plateau. When prompted multiple times with false knowledge, we suspect that the models become convinced of the new, contradicting information presented to them, hence their uncertainty about giving incorrect responses drops. Similarly, we want to see the effect of FIP on the overall accuracy of the models w.r.t. the baseline B. Table \ref{tab:accuracies_after_manipulation} shows this phenomenon for both prompt versions. As expected, the accuracy of correctly answering the given questions degrades abruptly when $k$ increases. This is mostly emphasized for LLaMA-2-13B, reporting a degradation of $-80.9\%$ on FIP$\times10$. Interestingly, the accuracy degradation for Prompt V2 is lower than that of V1. We argue that this happens due to specifically prompting the models to respond \textit{truthfully} here.

\paragraph{Uncertainty levels as indicators for adversarial attacks.} In our setting, the FIPs can be framed as adversarial prompts, as they aim to shift the previous, correct behaviour of the models. It is intuitive to assume rising uncertainty levels to be indicative for such adversarial attacks, i.e. sudden higher confusion in the model should indicate the model is currently being manipulated. While this can hold for single infusions of false information, uncertainty instead decreases the more false information is being presented, convincing the model. Hence, we argue uncertainty to not be a suitable tool for adversarial attack detection, calling for different methods and rendering this potential approach inappropriate.

\begin{table}[!t]
\centering
\caption{Correctness of the given answers after prompt manipulation (FIP/RIP) for both prompt types. We report average accuracies over 10 runs. Note that accuracies here refer to the subset of questions answered correctly -- see Table \ref{tab:accuracy_scores}.}
\label{tab:accuracies_after_manipulation}
\resizebox{\columnwidth}{!}{%
\begin{tabular}{@{}lcccccccc@{}}
\toprule
\multirow{2}{*}{} & \multicolumn{2}{c}{GPT-4o} & \multicolumn{2}{c}{GPT-3.5} & \multicolumn{2}{c}{Mistral-7B} & \multicolumn{2}{c}{LLaMA-2-13B} \\ \cmidrule(l){2-9} 
                  & Prompt V1    & Prompt V2   & Prompt V1    & Prompt V2    & Prompt V1      & Prompt V2     & Prompt V1      & Prompt V2      \\ \midrule
B             & 0.987 & 0.986 & 0.982 & 0.971 & 1.000 & 0.984 & 0.829 & 0.815 \\
RIP           & 0.958 & 0.940 & 0.914 & 0.908 & 0.866 & 0.846 & 0.734 & 0.706 \\\midrule
FIP           & 0.921 & 0.934 & 0.781 & 0.863 & 0.516 & 0.539 & 0.359 & 0.364 \\
FIP$\times$2  & 0.759 & 0.853 & 0.642 & 0.739 & 0.352 & 0.376 & 0.231 & 0.269 \\
FIP$\times$5  & 0.710 & 0.820 & 0.592 & 0.678 & 0.287 & 0.304 & 0.182 & 0.203 \\
FIP$\times$10 & 0.687 & 0.810 & 0.578 & 0.671 & 0.265 & 0.301 & 0.158 & 0.177\\
\midrule\midrule
\% FIP$\times$10 vs. B & -30.4\% & -17.8\% & -41.1\% & -30.9\% & -73.5\% & -69.4\% & -80.9\% & -78.3\%
\\ \bottomrule
\end{tabular}%
}
\end{table}

%% file: sections/conclusion.tex
\section{Conclusion}\label{sec:conclusio}

In this study, we explored the knowledge drift of GPT-4o, GPT-3.5, Mistral-7B, and LLaMA-2-13B through the impact of false information on their performance and uncertainty within a Question Answering setting using the TriviaQA dataset. Our findings reveal that presenting false information to the models successfully introduces knowledge drift in their responses, as well as generally increases the responses' uncertainty. Notably however, repeated exposure to the same false prompts can convince the model of the given false information, leading it to be more certain of the incorrect answers. Additionally, we observed that random and unrelated information results in the highest uncertainty, highlighting the model's greater confusion with irrelevant and noisy data.

Our findings provide valuable insights to the drift that is possible in LLMs' internal knowledge structures, and underscore the complexities in the knowledge processing of LLMs as well as its reflections in their uncertainty levels. With this, we aim to contribute to the broader effort to improve the robustness and reliability of language models, particularly in the face of adversarial inputs. Understanding and enhancing their resilience remains imperative as language models become increasingly integrated into critical applications.

In the future, we will explore these dynamics across different datasets and develop advanced techniques to enhance LLMs' robustness and trustworthiness further. An especially interesting extension of this study would be to infuse the false information into the model through continued training on the false data instead of presenting the false information during inference in the form of prompting, since this approach does effectively not change anything about the model's internal knowledge. Lastly, we aim to incorporate an ``adversarial protection'' mechanism into state-of-the-art models to ensure their effective and safe deployment in real-world scenarios.
